%% file: main.tex
\documentclass[runningheads,a4paper]{llncs}
\usepackage[T1]{fontenc}
\usepackage{graphicx}
\usepackage{amsmath,amsfonts,amssymb}
\usepackage{graphicx}
\usepackage[colorlinks=true, allcolors=blue]{hyperref}
\usepackage{times}
\usepackage{float}
\usepackage{booktabs}
\usepackage{array} 
\begin{document}
\title{Disentangling Heterogeneous Traffic Dynamics for Multi-Step Traffic Forecasting via Adaptive Spectral Decomposition}
\author{Zijun Huang\inst{1}  \and
Chenrui Fu\inst{1}  \and
Wenhao Wang\inst{1} \and
Xiaochuan Gou\inst{2} \and
Chih-Chieh Hung\inst{3} \and
Guanyao Li\inst{1}\thanks{Corresponding author: \email{guanyaoli@bnbu.edu.cn}}
}
\authorrunning{Z. Huang et al.}
\institute{Beijing Normal-Hong Kong Baptist University \\
\and Dalian Maritime University \and 
National Chung Hsing University
\\
}

\title{Disentangling Heterogeneous Traffic Dynamics for
Multi-Step Traffic Forecasting via Adaptive Spectral Decomposition}
\titlerunning{Adaptive Spectral Decomposition for Traffic Forecasting}

\maketitle%
\begin{abstract}
Accurate multi-step traffic forecasting remains challenging because observed traffic signals contain heterogeneous temporal dynamics with different characteristics and levels of predictability. Existing approaches typically model these dynamics within a unified representation or rely on predefined decomposition rules, which may limit their ability to flexibly separate persistent patterns from rapidly varying fluctuations. To address this issue, we propose the Adaptive Decomposition Network (ADNet), a component-specific forecasting framework that adaptively disentangles traffic dynamics into dominant and residual components. ADNet introduces a learnable complementary spectral decomposition mechanism that determines the contribution of each frequency bin to the two components. Unlike hard frequency partitioning, every frequency bin can contribute to both components with different learned proportions, allowing the decomposition to be optimized jointly with the forecasting objective. The reconstructed components are then modeled by two dedicated spatiotemporal forecasting branches, and their predictions are integrated to generate the final multi-step forecast. Experiments on the Alameda and Orange regions of the TraffiDent dataset show that ADNet achieves the best performance in 20 of the 24 reported region--horizon--metric comparisons, with particularly clear gains at longer forecasting horizons. Capacity-controlled ablation experiments further show that the learnable decomposition substantially outperforms a fixed decomposition and provides additional improvements beyond the dual-branch architecture alone. These results demonstrate the effectiveness of adaptive decomposition and component-specific modeling for multi-step traffic forecasting.

\end{abstract}
\input{1-Introduction}
\input{2-Related-Work}
\input{3-Methodology}

\input{4-Evaluation}
\input{5-Conclusion}

\end{document}

%% file: 1-Introduction.tex
\section{Introduction}

Accurate traffic forecasting is a fundamental capability of intelligent transportation systems and supports a wide range of applications, including traffic management, congestion mitigation, route planning, and transportation resource allocation. Given historical traffic observations collected from multiple locations in a road network, multi-step traffic forecasting aims to predict traffic conditions over multiple future time steps. Compared with one-step forecasting, multi-step forecasting provides a longer view of future traffic evolution, enabling transportation systems to anticipate upcoming conditions and make proactive decisions. However, accurately predicting traffic states over extended forecasting horizons remains challenging because traffic observations comprise heterogeneous and continuously evolving temporal dynamics with different forecasting characteristics.

A fundamental challenge is that these heterogeneous traffic dynamics are inherently entangled within observed traffic signals. Some variations are relatively stable and persistent, reflecting regular mobility behaviors and slowly evolving traffic states, whereas others exhibit rapid and irregular fluctuations caused by transient congestion, changing travel demand, incidents, and other dynamic factors. These dynamics not only exhibit different temporal characteristics but also possess different levels of predictability. Relatively stable patterns can often provide reliable information for forecasting over longer horizons, while rapidly varying fluctuations are more difficult to extrapolate and may introduce increasing uncertainty as the forecasting horizon extends. When such heterogeneous dynamics are modeled together within a unified representation, a forecasting model is required to simultaneously capture patterns with substantially different behaviors, potentially making the forecasting task unnecessarily difficult. This motivates explicitly disentangling heterogeneous traffic dynamics before forecasting.

Existing traffic forecasting studies have made substantial progress by developing increasingly powerful models for capturing complex spatial and temporal dependencies. Most approaches focus on improving the ability of a unified model to represent the observed traffic sequence~\cite{li2018dcrnn,yu2018stgcn,wu2019graph}, while another line of research explicitly decomposes traffic signals into different components before prediction~\cite{shao2022d2stgnn,fang2023stwave}. Although both directions have demonstrated promising performance, they still leave room for improvement in handling heterogeneous traffic dynamics. Unified modeling approaches may insufficiently distinguish dynamics with different forecasting characteristics, whereas decomposition-based approaches often rely on predefined assumptions or relatively rigid criteria to determine how different patterns should be separated. Consequently, an important question remains: how can heterogeneous traffic dynamics be explicitly disentangled while allowing the decomposition itself to adapt to the forecasting objective?

Addressing this question is challenging because heterogeneous traffic dynamics may not be separated by a clear and fixed boundary. Slowly varying patterns are generally more closely associated with persistent traffic dynamics, while rapidly changing variations are more likely to reflect irregular fluctuations. Nevertheless, this distinction is not absolute. Information at the same temporal frequency may contribute to different traffic dynamics to different degrees. Assigning a particular frequency exclusively to one component may therefore impose an overly restrictive representation and discard potentially useful information. Instead of determining the decomposition through a predefined frequency boundary, it is desirable to learn how different spectral patterns should contribute to different traffic dynamics directly from data.

Motivated by this observation, we propose the \textbf{Adaptive Decomposition Network (ADNet)}\footnote{Code is available at {\urlstyle{same}\url{https://github.com/xiaohuliming/Xtraffic-forecast}}.}, a multi-step traffic forecasting framework that adaptively disentangles heterogeneous traffic dynamics and models the resulting components separately. ADNet decomposes historical traffic observations into a dominant component and a residual component. The dominant component is intended to emphasize relatively persistent and predictable traffic dynamics, whereas the residual component focuses on irregular deviations and rapidly changing variations. Rather than imposing a hard separation between the two components, ADNet learns how traffic patterns at different frequencies should be distributed between them. In this way, the decomposition adapts to the intrinsic characteristics of traffic data and the forecasting objective instead of relying on manually defined frequency boundaries.

Specifically, ADNet transforms historical traffic observations into the frequency domain and introduces a learnable complementary spectral decomposition mechanism. For each frequency bin, the model learns how its information should be distributed between the dominant and residual components through complementary weights. Therefore, every frequency bin can contribute to both components with different learned proportions, allowing the two components to exploit information from the entire frequency spectrum while emphasizing different temporal dynamics. The decomposed spectra are then transformed back into the time domain and modeled through two dedicated forecasting branches. Finally, the predictions from the two branches are integrated to generate the multi-step traffic forecasts. Through this design, ADNet combines adaptive decomposition with component-specific forecasting to explicitly model heterogeneous traffic dynamics.

The main contributions of this work are summarized as follows:
\begin{itemize}

\item \textbf{Component-Specific Forecasting Framework.} We formulate multi-step traffic forecasting from the perspective of disentangling heterogeneous traffic dynamics and propose ADNet, a component-specific forecasting framework that explicitly decomposes traffic observations into dominant and residual components. The two components are modeled through dedicated forecasting branches, enabling traffic dynamics with different temporal characteristics and levels of predictability to be captured separately.

\item \textbf{Learnable Complementary Spectral Decomposition.} We introduce a learnable complementary spectral decomposition mechanism that adaptively determines how different frequency patterns contribute to the dominant and residual components. Unlike predefined frequency boundaries or hard frequency selection, the proposed mechanism allows every frequency bin to contribute to both components with complementary learned proportions, providing a flexible decomposition of heterogeneous traffic dynamics.

\item \textbf{Empirical Validation and Ablation.} Extensive experiments on real-world traffic data demonstrate the effectiveness of ADNet across different forecasting horizons and evaluation metrics. Capacity-controlled ablation experiments further investigate the contributions of the dual-branch architecture and adaptive decomposition mechanism, showing that learning the decomposition is important for improving multi-step traffic forecasting performance.

\end{itemize}

%% file: 2-Related-Work.tex
\section{Related Work}

\subsection{Spatiotemporal Traffic Forecasting}

Spatiotemporal modeling has become a dominant paradigm for traffic forecasting because traffic observations exhibit strong dependencies across both road-network locations and time. Representative methods include DCRNN, which integrates diffusion graph convolution with recurrent modeling, STGCN, which combines graph and temporal convolutions, and Graph WaveNet (GWN), which introduces dilated causal convolutions and adaptive graph learning~\cite{li2018dcrnn,yu2018stgcn,wu2019graph}. More recent studies further explore adaptive and dynamic spatial dependencies. For example, D$^2$STGNN decomposes traffic signals into diffusion and inherent components and models them together with dynamic graph learning~\cite{shao2022d2stgnn}.

These methods have substantially improved the modeling of complex spatiotemporal dependencies. Their primary focus, however, is on designing more expressive spatial or temporal representations. Our work addresses a complementary problem: how heterogeneous temporal dynamics within traffic observations can be disentangled before forecasting. ADNet decomposes the input into different temporal components and models them through separate forecasting branches, allowing component-specific modeling to complement existing spatiotemporal forecasting architectures.

\subsection{Decomposition-Based Time-Series Forecasting}

Time-series decomposition provides a natural way to model signals containing heterogeneous temporal patterns. Classical methods such as STL separate observations into trend, seasonal, and remainder components~\cite{cleveland1990stl}. More recently, decomposition has been incorporated into deep forecasting models. Autoformer progressively separates trend and seasonal information~\cite{wu2021autoformer}, FEDformer combines series decomposition with frequency-enhanced modeling~\cite{zhou2022fedformer}, and DLinear demonstrates the effectiveness of explicitly modeling decomposed components with simple forecasting architectures~\cite{zeng2023dlinear}. MSD-Mixer further introduces multi-scale decomposition to capture temporal patterns at different scales~\cite{zhong2024msdmixer}.

These studies demonstrate that separating heterogeneous temporal patterns can facilitate forecasting. However, the resulting components are typically defined according to a particular decomposition formulation, such as trend--seasonality or multi-scale structures. In contrast, ADNet learns how heterogeneous traffic dynamics should be decomposed according to their spectral contributions and the forecasting objective.

\subsection{Frequency-Aware Traffic Forecasting}

Frequency-domain modeling provides another perspective for characterizing heterogeneous temporal dynamics. StemGNN exploits graph and temporal spectral representations for multivariate forecasting, while FreTS models intra-series and inter-series dependencies through frequency-domain transformations~\cite{cao2020stemgnn,yi2023frets}. FreqMoE further decomposes time-series representations into frequency components and employs specialized experts to capture different frequency characteristics~\cite{liu2025freqmoe}.

Frequency-aware modeling has also been increasingly explored in traffic forecasting. STWave employs wavelet decomposition to separate relatively stable trends from fluctuating traffic variations~\cite{fang2023stwave}. DFDGCN exploits Fourier representations to alleviate temporal-shift effects when learning dynamic spatial dependencies~\cite{li2024dfdgcn}. LHFNet explicitly models low- and high-frequency traffic characteristics using different encoders~\cite{feng2025lhfnet}, while HyperD decomposes traffic dynamics into periodic and residual components and models them through specialized mechanisms~\cite{shao2026hyperd}.

ADNet shares the general motivation of exploiting heterogeneous frequency characteristics but differs in how the decomposition is constructed. Rather than assigning spectral information exclusively to predefined components or frequency ranges, ADNet learns complementary weights that determine the contribution of each frequency bin to both the dominant and residual components. Consequently, every frequency can contribute to both components with different learned proportions. This allows the decomposition itself to be optimized jointly with the forecasting objective, providing a more flexible way to disentangle heterogeneous traffic dynamics without imposing a hard frequency partition.

%% file: 3-Methodology.tex
\section{Methodology}
\label{sec:methodology}

\subsection{Model Overview}
\label{sec:overall-framework}

\begin{figure}[t]
    \centering
    \includegraphics[width=0.55\linewidth]{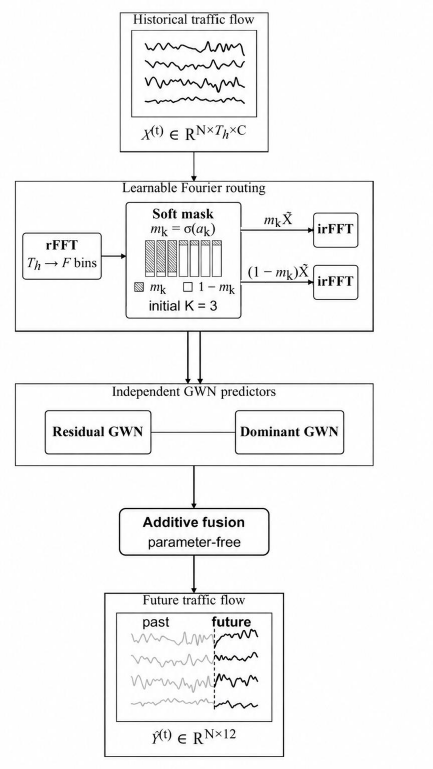}
    \caption{Overview of Adaptive Decomposition Network (ADNet).}
    \label{fig:placeholder}
\end{figure}

Figure~\ref{fig:placeholder} illustrates the overall architecture of the proposed Adaptive Decomposition Network (ADNet). Given historical traffic observations from multiple locations, ADNet first transforms the input sequences into the frequency domain using the real-valued fast Fourier transform (rFFT). A learnable complementary spectral decomposition module then adaptively distributes the information in each frequency bin between a dominant component and a residual component. Unlike hard frequency partitioning, every frequency bin can contribute to both components with different learned proportions, allowing the decomposition to adapt to the forecasting objective.

The two decomposed spectra are subsequently transformed back into the time domain through inverse rFFT (irFFT) and modeled by two independently parameterized Graph WaveNet (GWN) branches~\cite{wu2019graph}. The two branches capture the spatiotemporal dependencies of the dominant and residual dynamics separately and produce their respective multi-step forecasts. Finally, the two predictions are combined through element-wise addition to obtain the final traffic-flow prediction. In this way, ADNet integrates adaptive spectral decomposition with component-specific spatiotemporal forecasting to explicitly model heterogeneous traffic dynamics.

\subsection{Learnable Complementary Spectral Decomposition}

Given historical traffic observations
$\mathbf{X}\in\mathbb{R}^{N\times T_h\times C}$,
where $N$ denotes the number of locations, $T_h$ the historical window length,
and $C$ the number of input features, we first transform the traffic sequences
into the frequency domain using the real Fast Fourier Transform (rFFT) along
the temporal dimension:
\begin{equation}
    \widetilde{\mathbf{X}}
    =\operatorname{rFFT}(\mathbf{X}),
    \qquad
    \widetilde{\mathbf{X}}\in\mathbb{C}^{N\times F\times C},
\end{equation}
where $F=\lfloor T_h/2\rfloor+1$ is the number of non-redundant frequency bins.
The transformation is independently applied to each location and input
channel.

Instead of using a fixed frequency cutoff to separate different traffic
dynamics, we introduce a learnable soft mask
$\mathbf{m}=[m_0,\ldots,m_{F-1}]$ to determine the contribution of each
frequency bin to the dominant component. For the $k$-th frequency bin, its
weight is defined as
\begin{equation}
    m_k=\sigma(a_k), \qquad m_k\in(0,1),
\end{equation}
where $a_k$ is a learnable parameter and $\sigma(\cdot)$ denotes the sigmoid
function. The mask is shared across samples, locations, and input channels and
is optimized jointly with the forecasting model.

To provide an inductive bias at the beginning of training, we initialize the
mask such that the first $K$ frequency bins receive larger dominant weights:
\begin{equation}
    a_k^{(0)}=
    \begin{cases}
        \alpha, & k<K,\\
        -\alpha, & k\geq K,
    \end{cases}
\end{equation}
where $K$ is a hyperparameter satisfying $1\leq K\leq F$, and $\alpha>0$
controls the strength of the initialization bias. Importantly, $K$ only
determines the initialization and does not impose a hard frequency boundary.
All mask parameters remain trainable and may adapt freely during optimization.

The dominant and residual spectra are constructed using complementary weights:
\begin{equation}
    \widetilde{\mathbf{X}}_{\mathrm{D}}
    =\mathbf{m}\odot\widetilde{\mathbf{X}},
\end{equation}

\begin{equation}
    \widetilde{\mathbf{X}}_{\mathrm{R}}
    =(\mathbf{1}-\mathbf{m})\odot\widetilde{\mathbf{X}},
\end{equation}
where $\odot$ denotes element-wise multiplication along the frequency
dimension. Therefore, each frequency bin can contribute to both components
with different proportions rather than being assigned exclusively to one of
them.

Finally, the two spectra are transformed back into the time domain through
irFFT:
\begin{equation}
    \mathbf{X}_{\mathrm{D}}
    =\operatorname{irFFT}(\widetilde{\mathbf{X}}_{\mathrm{D}}),
\end{equation}
\begin{equation}
    \mathbf{X}_{\mathrm{R}}
    =\operatorname{irFFT}(\widetilde{\mathbf{X}}_{\mathrm{R}}).
\end{equation}
Because the two spectral masks are complementary, the reconstructed sequences
satisfy
$\mathbf{X}_{\mathrm{D}}+\mathbf{X}_{\mathrm{R}}=\mathbf{X}$
up to numerical precision. The proposed decomposition therefore preserves the
complete historical signal while allowing the model to learn different
spectral emphasis for the dominant and residual traffic dynamics.

The two reconstructed components are subsequently processed by two parallel and independently parameterized Graph WaveNet (GWN) branches~\cite{wu2019graph}. The dominant branch learns the spatiotemporal dependencies associated with relatively persistent and regular traffic dynamics, whereas the residual branch focuses on the spatial and temporal correlations underlying irregular and rapidly varying traffic fluctuations. Although the two branches adopt the same GWN backbone, their parameters are learned independently, enabling them to specialize in the distinct characteristics of the two components. Each branch directly generates multi-step predictions in the time domain, resulting in a dominant-component forecast and a residual-component forecast.

\subsection{Component-Specific Spatiotemporal Forecasting}

After spectral decomposition, the reconstructed dominant and residual sequences exhibit different temporal characteristics and are therefore modeled separately. We employ two independently parameterized spatiotemporal forecasting branches, one for each component, allowing the predictors to specialize in the distinct dynamics contained in the dominant and residual signals.

In this work, Graph WaveNet (GWN)~\cite{wu2019graph} is adopted as the forecasting backbone because of its effectiveness in jointly capturing temporal dependencies and spatial correlations among traffic locations. The dominant and residual sequences are independently fed into two GWN branches to produce their corresponding multi-step forecasts. Although the two branches share the same network architecture, their parameters are not shared, enabling each branch to learn component-specific spatiotemporal representations.

In this work, GWN is used as the forecasting backbone for both branches. While the proposed component-specific modeling strategy is not inherently dependent on GWN, evaluating alternative spatiotemporal backbones is beyond the scope of this study.

\subsection{Prediction Fusion and Learning Objective}
\label{sec:learning-objective}

The dominant and residual branches independently generate multi-step forecasts,
denoted by $\widehat{\mathbf{Y}}_{\mathrm{D}}$ and
$\widehat{\mathbf{Y}}_{\mathrm{R}}$, respectively. We combine the two
predictions through element-wise addition:
\begin{equation}
    \widehat{\mathbf{Y}}
    =
    \widehat{\mathbf{Y}}_{\mathrm{D}}
    +
    \widehat{\mathbf{Y}}_{\mathrm{R}},
\end{equation}
where $\widehat{\mathbf{Y}}$ denotes the final multi-step traffic forecast.
The additive fusion introduces no additional parameters and naturally combines
the complementary information learned from the two components.

The entire framework is trained end-to-end using the mean absolute error
(MAE) between the predicted and ground-truth traffic values:
\begin{equation}
    \mathcal{L}
    =
    \frac{1}{B\times N\times T_p}
    \sum_{b=1}^{B}
    \sum_{n=1}^{N}
    \sum_{h=1}^{T_p}
    \left|
    \widehat{Y}_{b,n,h}
    -
    Y_{b,n,h}
    \right|,
\end{equation}
where $B$, $N$, and $T_p$ denote the batch size, number of traffic locations,
and prediction horizon, respectively. The forecasting loss jointly optimizes
the learnable spectral decomposition module and the two GWN forecasting
branches.

%% file: 4-Evaluation.tex
\section{Evaluation}

\subsection{Datasets}
We conduct experiments on the TraffiDent benchmark~\cite{gou2024xtraffic}, which provides large-scale traffic data collected across California in 2023. To focus on freeway mainline traffic dynamics, we retain only mainline sensors and aggregate the traffic measurements into 5-minute intervals. We construct two regional datasets corresponding to Alameda and Orange counties. Their statistics are summarized in Table~\ref{tab:dataset}.

Each region's time series is chronologically divided into training, validation, and test periods in a 70\%/15\%/15\% ratio. Forecasting windows are sampled around incidents, deduplicated by forecast origin, and filtered to prevent prediction targets or incidents from being shared across splits.

\begin{table}[t]
\centering
\caption{Spatial statistics and temporal coverage of the Alameda and Orange datasets.}
\label{tab:dataset}
\begin{tabular}{lrrl}
\toprule
Region & Nodes & Edges & Time range \\
\midrule
Alameda & 521 & 13,828 & 2023-01-01 to 2023-12-31 \\
Orange  & 990 & 29,142 & 2023-01-01 to 2023-12-31 \\
\bottomrule
\end{tabular}
\end{table}

\subsection{Baselines}
We compare ADNet with ten representative traffic forecasting methods. Historical Last (HL) and LSTM are included as non-graph temporal baselines. Moreover, DCRNN~\cite{li2018dcrnn}, STGCN~\cite{yu2018stgcn}, and Graph WaveNet (GWN)~\cite{wu2019graph} are representative spatiotemporal graph forecasting models, with GWN serving as the direct backbone baseline of ADNet. We further compare with AGCRN~\cite{bai2020agcrn}, which learns adaptive spatial dependencies, ASTGCN~\cite{guo2019astgcn}, which incorporates spatial--temporal attention, and more recent graph-based approaches including DSTAGNN~\cite{lan2022dstagnn} and D$^2$STGNN~\cite{shao2022d2stgnn}. STWave~\cite{fang2023stwave} uses wavelet decomposition to separate traffic signals into stable trends and fluctuating events, which are modeled by a dual-channel spatiotemporal network with efficient spectral graph attention. This provides a decomposition-based comparison for ADNet. Together, these baselines cover temporal, graph-based, attention-based, adaptive, and dynamic spatiotemporal forecasting strategies.

\subsection{Implementation Details}
All methods use the same data splits and forecasting protocol. The input window contains 12 historical time steps, and the model predicts the next 12 time steps; at 5-minute intervals, these correspond to one hour of history and one hour of future traffic flow. For the baseline methods, we follow their recommended model and training configurations whenever applicable. In the main comparison, all trainable methods are evaluated using seed 42 and trained for at most 30 epochs, with model selection based on validation performance.

For ADNet, each forecasting branch adopts GWN with 32 hidden channels and a dropout rate of 0.3. The model is optimized using Adam with an initial learning rate of $10^{-3}$ and a weight decay of $10^{-4}$. For the adaptive decomposition module, the complementary spectral weights are initialized using a boundary of $K=3$ and an initialization strength of $\alpha=2$, unless otherwise specified. These parameters affect only the initialization of the spectral weights; all frequency weights are subsequently optimized jointly with the forecasting model.

\subsection{Evaluation Metrics}
We evaluate forecasting accuracy using three widely adopted metrics: mean absolute error (MAE), root mean squared error (RMSE), and mean absolute percentage error (MAPE), where lower values indicate better performance. All metrics are computed on the original traffic-flow scale. We report performance at H3, H6, and H12, corresponding to 15-, 30-, and 60-minute-ahead forecasting, respectively. We additionally report \emph{Average}, which aggregates predictions over all 12 forecasting horizons.

Missing target observations are excluded using the original observation mask. For MAPE, target values whose absolute flow is close to zero are additionally excluded to avoid numerical instability.

\subsection{Overall Performance}

\input{4-Evaluation-FullTable}

Table~\ref{tab:main-results} reports the forecasting performance of ADNet and the compared methods on the Alameda and Orange regions of TraffiDent. We report MAE, RMSE, and MAPE, where lower values indicate better forecasting performance. The best and second-best results in each setting are highlighted in bold and underlined, respectively.

Overall, ADNet achieves consistently strong performance across both regions, forecasting horizons, and evaluation metrics. Among the 24 region--horizon--metric combinations reported in Table~\ref{tab:main-results}, ADNet obtains the best result in 20 cases and the second-best result in the remaining four cases. This broad improvement across MAE, RMSE, and MAPE indicates that the advantage of ADNet is not restricted to a particular forecasting horizon or error metric.

On Alameda, ADNet achieves the best Average MAE and RMSE of 12.13 and 22.32, respectively. ADNet also achieves the best MAE and RMSE at H6 and the best performance across all three metrics at H12. At H3, it obtains the lowest MAE, while its RMSE of 20.00 is nearly identical to the best result of 19.99 achieved by D$^2$STGNN.

The advantage of ADNet is particularly consistent on Orange, where it achieves the best result for every metric at every reported horizon. For the Average performance, ADNet reduces MAE, RMSE, and MAPE by 2.79\%, 1.92\%, and 2.83\%, respectively, compared with the strongest baseline results. Moreover, its advantage becomes more pronounced at longer forecasting horizons. For example, relative to GWN, the MAPE reduction increases from 2.16\% at H3 to 2.62\% at H6 and 3.69\% at H12. Similar improvements are observed for MAE and RMSE at the longer horizons.

STWave provides a comparison with predefined wavelet decomposition. ADNet reduces Average MAE relative to STWave by 1.55\% on Alameda and 5.10\% on Orange, and reduces Average RMSE by 2.85\% and 6.88\%, respectively. The MAPE comparison is more mixed: on Alameda, STWave achieves lower MAPE at H3, H6, and Average, with values of 15.92\%, 17.49\%, and 17.82\%, compared with 16.29\%, 17.78\%, and 17.91\% for ADNet. ADNet achieves lower MAPE at H12 on Alameda and at all reported horizons on Orange.

These results demonstrate the effectiveness of ADNet for multi-step traffic forecasting. In particular, the consistent gains at longer forecasting horizons are aligned with the motivation of adaptive decomposition: disentangling heterogeneous traffic dynamics allows components with different temporal characteristics to be modeled separately before their predictions are integrated. We further examine the contributions of the dual-branch architecture and the adaptive decomposition mechanism through ablation experiments.

\subsection{Ablation Study}

\begin{table}[t]
\centering
\caption{Ablation study on Alameda.}
\label{tab:xtraffic-ablation-metrics}
\begingroup
\small
\setlength{\tabcolsep}{7pt}
\renewcommand{\arraystretch}{1.12}
\begin{tabular}{@{}lrrr@{}}
\toprule
Model & MAE $\pm$ SD & RMSE $\pm$ SD & MAPE \% $\pm$ SD \\
\midrule
\multicolumn{4}{l}{\textbf{Alameda}} \\
GWN & $11.0620 \pm 0.1059$ & $20.9633 \pm 0.1582$ & $19.4566 \pm 0.1986$ \\
DualGWN & $10.7346 \pm 0.0518$ & $20.4777 \pm 0.1319$ & $\mathbf{18.7431} \pm 0.1477$ \\
ADNet fixed-K3 & $11.1033 \pm 0.1101$ & $20.9241 \pm 0.1138$ & $19.4989 \pm 0.0615$ \\
ADNet learnable-K3 & $\mathbf{10.7146} \pm 0.0777$ & $\mathbf{20.3995} \pm 0.1042$ & $18.8070 \pm 0.1510$ \\
\bottomrule
\end{tabular}
\endgroup
\end{table}

We conduct an ablation study to investigate two questions: whether the improvement of ADNet can be explained by the additional capacity of the dual-branch architecture, and whether learning the decomposition provides an advantage over a fixed spectral partition. All variants are trained for 100 epochs using three random seeds, with the checkpoint selected according to validation MAE. Table~\ref{tab:xtraffic-ablation-metrics} reports the mean and standard deviation of the Average performance over all 12 forecasting horizons.

We first evaluate the effect of the dual-branch architecture. Compared with the single-branch GWN, DualGWN reduces MAE, RMSE, and MAPE by 2.96\%, 2.32\%, and 3.67\%, respectively. This result shows that using two independently parameterized forecasting branches already provides a clear performance benefit and therefore serves as an important capacity-matched control for evaluating the decomposition mechanism.

We next examine whether the decomposition strategy itself matters. ADNet with a fixed decomposition does not improve over GWN and produces slightly higher MAE and MAPE. In contrast, learning the decomposition substantially improves performance. Compared with the fixed variant, ADNet with learnable decomposition reduces MAE, RMSE, and MAPE by 3.50\%, 2.51\%, and 3.55\%, respectively. This result indicates that simply separating the signal into components is insufficient; the decomposition needs to adapt to the forecasting objective.

Finally, compared with the capacity-matched DualGWN, the learnable ADNet variant further reduces MAE from 10.7346 to 10.7146 and RMSE from 20.4777 to 20.3995, while achieving a comparable MAPE (18.8070 versus 18.7431). These results suggest that the performance gain of ADNet cannot be attributed solely to increased model capacity. Rather, the learnable decomposition provides additional benefit by adaptively organizing heterogeneous traffic dynamics before component-specific forecasting.

%% file: 4-Evaluation-FullTable.tex
\begin{table}[t]
\centering
\caption{Performance comparison on the Alameda and Orange regions of TraffiDent. H3, H6, and H12 denote the 15-, 30-, and 60-minute forecasting horizons, respectively, while Average aggregates all 12 horizons.}
\label{tab:main-results}
\begingroup
\setlength{\tabcolsep}{1.6pt}
\renewcommand{\arraystretch}{0.86}
\resizebox{\textwidth}{!}{%
\begin{tabular}{l|rrr|rrr|rrr|rrr}
\hline
Method & \multicolumn{3}{c|}{H3} & \multicolumn{3}{c|}{H6} & \multicolumn{3}{c|}{H12} & \multicolumn{3}{c}{Average} \\
 & MAE & RMSE & MAPE (\%) & MAE & RMSE & MAPE (\%) & MAE & RMSE & MAPE (\%) & MAE & RMSE & MAPE (\%) \\
\hline
\multicolumn{13}{l}{\textbf{Alameda County}} \\
\hline
HL & 14.57 & 25.23 & 20.87 & 17.67 & 30.23 & 24.57 & 24.18 & 40.62 & 33.00 & 18.26 & 31.76 & 25.46 \\
LSTM & 11.48 & 20.82 & 16.43 & 12.86 & 23.47 & 18.21 & 15.29 & 27.79 & 21.51 & 12.96 & 23.76 & 18.44 \\
DCRNN & 11.37 & 20.43 & 16.67 & 12.60 & 22.74 & 18.61 & 14.77 & 26.23 & 22.04 & 12.68 & 22.84 & 18.82 \\
AGCRN & 11.65 & 22.64 & 31.96 & 12.62 & 24.96 & 35.40 &  \underline{13.98} & 28.00 & 39.05 & 12.57 & 24.83 & 34.47 \\
STGCN & 12.58 & 22.87 & 32.21 & 13.32 & 24.37 & 32.80 & 14.72 & 26.88 & 34.30 & 13.41 & 24.53 & 33.01 \\
GWN & 11.28 & 20.33 & 16.50 & 12.53 & 22.68 & 18.14 & 14.69 & 26.59 & 21.15 & 12.62 & 22.99 & 18.34 \\
ASTGCN & 12.15 & 21.12 & 19.96 & 13.57 & 23.58 & 22.57 & 15.79 & 27.34 & 27.14 & 13.55 & 23.72 & 22.54 \\
DSTAGNN & 11.70 & 20.67 & 18.72 & 12.45 & \underline{22.39} & 19.03 & 14.19 & \underline{25.65} & 21.41 & 12.55 & \underline{22.64} & 19.25 \\
D$^2$STGNN & 11.22 & \textbf{19.99} & 18.12 & 12.92 & 22.71 & 19.77 & 14.99 & 26.27 & 22.32 & 13.02 & 23.01 & 19.85 \\
STWave & \underline{11.03} & 20.29 & \textbf{15.92} & \underline{12.22} & 22.72 & \textbf{17.49} & 14.31 & 26.55 & \underline{20.58} & \underline{12.32} & 22.98 & \textbf{17.82} \\
\textbf{ADNet (ours)} & \textbf{11.01} & \underline{20.00} & \underline{16.29} & \textbf{12.10} & \textbf{22.11} & \underline{17.78} & \textbf{13.86} & \textbf{25.48} & \textbf{20.41} & \textbf{12.13} & \textbf{22.32} & \underline{17.91} \\
\hline
\multicolumn{13}{l}{\textbf{Orange County}} \\
\hline
HL & 15.41 & 26.21 & 20.29 & 18.47 & 31.08 & 23.42 & 24.89 & 40.99 & 30.99 & 19.03 & 32.38 & 24.22 \\
LSTM & 12.25 & 21.85 & 16.00 & 13.72 & 24.59 & 17.57 & 16.32 & 29.16 & 20.48 & 13.81 & 24.86 & 17.76 \\
DCRNN & 14.56 & 23.66 & 26.27 & 17.85 & 28.55 & 30.72 & 25.49 & 39.41 & 40.16 & 18.63 & 30.22 & 31.30 \\
AGCRN & 12.82 & 26.23 & 37.02 & 13.74 & 28.54 & 40.00 & \underline{14.66} & 28.95 & 38.42 & 13.64 & 27.97 & 38.80 \\
STGCN & 13.36 & 25.19 & 31.00 & 14.06 & 26.46 & 31.27 & 15.24 & 28.40 & 32.15 & 14.07 & 26.48 & 31.41 \\
GWN & \underline{11.75} & \underline{20.81} & \underline{15.74} & \underline{12.89} & \underline{22.85} & \underline{16.82} & 14.77 & \underline{26.06} & \underline{18.97} & \underline{12.92} & \underline{22.95} & \underline{16.95} \\
ASTGCN & 13.09 & 22.50 & 18.68 & 14.75 & 25.50 & 20.66 & 17.97 & 30.97 & 23.25 & 14.91 & 25.96 & 20.17 \\
DSTAGNN & 12.13 & 21.12 & 17.76 & 13.27 & 23.13 & 20.42 & 15.31 & 26.66 & 26.04 & 13.36 & 23.32 & 20.81 \\
D$^2$STGNN & 12.23 & 21.15 & 21.47 & 13.73 & 23.42 & 23.76 & 16.05 & 26.98 & 26.46 & 13.79 & 23.54 & 24.50 \\
STWave & 11.88 & 21.39 & 15.88 & 13.19 & 23.99 & 17.07 & 15.46 & 28.14 & 19.57 & 13.24 & 24.17 & 17.14 \\
\textbf{ADNet (ours)} & \textbf{11.53} & \textbf{20.56} & \textbf{15.40} & \textbf{12.55} & \textbf{22.44} & \textbf{16.38} & \textbf{14.20} & \textbf{25.38} & \textbf{18.27} & \textbf{12.56} & \textbf{22.51} & \textbf{16.47} \\

\hline
\end{tabular}%
}
\endgroup
\end{table}

%% file: 5-Conclusion.tex
\section{Conclusion}

In this work, we proposed the Adaptive Decomposition Network (ADNet) for multi-step traffic forecasting. ADNet addresses the challenge of entangled heterogeneous traffic dynamics by adaptively decomposing historical observations into dominant and residual components and modeling them through separate forecasting branches. A learnable complementary spectral decomposition mechanism allows each frequency bin to contribute to both components with different learned proportions, avoiding the restriction of a hard frequency partition.

Experiments on the Alameda and Orange regions of TraffiDent demonstrate that ADNet consistently achieves strong performance across different forecasting horizons and evaluation metrics, with particularly clear improvements at longer horizons. The capacity-controlled ablation study further shows that, although the dual-branch architecture itself provides performance gains, learning the decomposition substantially outperforms a fixed decomposition and provides additional improvements beyond the capacity-matched control. These results demonstrate the effectiveness of combining adaptive decomposition with component-specific forecasting for modeling heterogeneous traffic dynamics.